\documentclass[letterpaper]{article}
\usepackage{aaai}
\usepackage{times}
\usepackage{helvet}
\usepackage{courier}
\usepackage{nicefrac}
\usepackage{xr}

\usepackage{times}
\usepackage{color}
\usepackage{latexsym}

\usepackage{graphicx}
\usepackage[centertags,fleqn]{amsmath}
\usepackage{amssymb,amsfonts,xspace}
\usepackage{booktabs}
\usepackage{multirow}
\newcommand{\highlight}[1]{{#1}}
\newcommand{\morehighlight}[1]{#1}

\title{Turing's Red Flag}

\author{Toby Walsh\\ University of New South Wales
and NICTA\\ Sydney, Australia}

\begin{document}
\maketitle

Movies can be a good place to see what the future looks like.
According to Robert Wallace, a retired director of the CIA's 
Office of Technical Service,

\begin{quote}
{\em   ``\ldots  When a new Bond movie was released, we always got calls asking, "Do you have one of those?" If I answered "no," the next question was, "How long will it take you to make it?" Folks didn't care about the laws of physics or that Q was an actor in a fictional series -- his character and inventiveness pushed our imagination \ldots'' } 
\cite{spycraft}
\end{quote}

As an example, 
the CIA successfully copied
the shoe-mounted spring-loaded and poison-tipped knife 
in ``From Russia With Love''. 
It's interesting to speculate
on what else Bond movies may have
led to being invented. 

For this reason, I've been 
considering what movies predict
about the future of Artificial Intelligence (AI).
One theme that emerges in several science fiction
movies is that of an AI
mistaken for human. 
In the classic movie {\em Blade Runner}, 
Rick Deckard (Harrison Ford) tracks down
and destroys replicants that have escaped 
and are visually indistinguishable from humans. 
Tantalisingly the film leaves open the question
of whether Rick Deckard is himself a replicant.
More recently, the movie {\em Ex Machina} centres 
around a type of Turing test in 
which the robot Ava tries
to be convincingly human enough to trick someone 
into helping her escape.
And in {\em Metropolis}, one of the very first 
science fiction movies ever, a robot
disguises itself as the woman Maria and
thereby causes the workers to revolt. 

It thus seems likely that 
sometime in the future we will have
to deal with the impact of AI's 
being mistaken for humans. 
In fact, it could be argued that this 
future is already here.
Joseph Weizenbaum proposed
ELIZA as a ``parody'' of a psychotherapist
and first described the program 
in the pages of {\em Communications of the ACM}
back in 1966 \cite{eliza}. 
However, his secretary famously asked 
to be left alone so she could talk in private 
to the chatterbot.
More recently a number of different
chatterbots have fooled judges in the 
annual Loebner prize, a 
somewhat crippled version of the Turing Test. 

Alan Turing, one of the father's of artificial
intelligence predicted in 1950 that computers would
be mistaken for humans in around fifty years
\cite{turingtest}. We may be running a little 
late on this prediction. Nevertheless the
test that Alan Turing proposed in
the very same paper that contains his fifty
year prediction remains 
the best known test for artificial intelligence
(even if there are efforts underway
to update and refine his
test). 
Let us not forget that
the Turing Test is all about
an artificial intelligence passing itself off as a human. 
Even if you are not a fan of the Turing Test,
it nevertheless has placed
the idea of computers emulating humans
firmly in our consciousness. 

As any lover of Shakespeare knows, there are 
many dangers awaiting us when we try
to disguise our identity. 
What happens if the AI impersonates
someone we trust? Perhaps they will be able
to trick us to do their bidding.
What if we suppose they have human level
capabilities but they can only act
at a sub-human level? Accidents might
quickly follow. What happens if we develop a 
social attachment to the AI? 
Or worse still, what if we fall 
in love with them? 
There is
a minefield of problems awaiting
us here.

This is not the first time in history
that a technology has come along
that might disrupt and endanger our lives.
Concerned about the impact of motor vehicles
on public safety, the UK parliament
passed the Locomotive Act in 1865.
This required a person to walk
in front of any motorised vehicle
with a red flag to signal the 
oncoming danger. Of course, public safety wasn't
the only motivation for this law
as the railways profited from 
restricting motor vehicles in this
way. Indeed, the law clearly
restricted the use of
motor vehicles to a greater
extent than safety alone required. \highlight{And 
this was a bad thing.} Nevertheless, the
sentiment was a good one: till 
society had adjusted to the arrival
of a new technology, the public
had a right to be forewarned of potential
dangers. 

Interestingly, this red flag law was withdrawn
three decades later in 1896 when the speed limit 
was raised to
14 mph (approximately 23 kmph). 
Coincidently the first speeding offence,
as well as the first British motoring
fatality, the unlucky pedestrian Bridget Driscoll
also occurred in that same year. And road accidents
have quickly escalated from then on. 
By 1926, the first year in which records are
available, there were 134,000 cases of serious injury,
yet there were only 1,715,421 vehicles on the roads of Great Britain.
That's one serious injury each year for every
13 vehicles on the road. And a century later, thousands still
die on our roads every year. 

Inspired by such historical precedents, I 
propose that a law be enacted to 
prevent AI systems from being mistaken
for humans. In recognition of Alan
Turing's seminal contributions to
this area, I am calling this the Turing Red Flag
law. 

~ \\

\highlight{
\fbox{
\begin{minipage}[t]{0.42\textwidth}
{\bf Turing Red Flag law:} An autonomous system should be
designed so that it is unlikely to be mistaken
for anything besides an autonomous sysem, 
and should identify itself
at the start of any interaction
with another agent.
\end{minipage}}
}

~ \\

\highlight{
Let me be clear. This is not the law itself but
a summary of its intent. Any law will
have to be much longer and much more
precise in its scope. Legal experts
as well as technologists 
will be needed to draft such a law.
The actual wording will need to be carefully
crafted, and the terms properly defined. 
\morehighlight{It will, for instance, require
a precise definition of autonomous
system. For now, we will consider any
system that has some sort of freedom to act
independently. Think, for instance, self-driving car. 
Though such a car does not choose its end destination,
it nevertheless does independently decide on the 
actual way to reach that given end destination. }
I would also expect that, as is often
the case in such matters, the exact definitions
will be left to the last moment to leave 
bargaining room to get any law into force. 
}

There are two parts to this proposed law.
\highlight{
The first part of the law states that an autonomous
system should not be designed to act in a way that 
it is likely to be mistaken there is a human 
in the loop. Of course, it is not impossible
to think of some situations where
it might be beneficial for an autonomous system
to be mistaken for something other than an
autonomous system.} An AI system pretending to be
human might, for example,
create more engaging interactive
fiction. 
More controversially, robots pretending to
be human might make better \morehighlight{care-givers} and companions
for the elderly. 
However, there are many more reasons
we don't want computers to be intentionally or
unintentionally fooling us. \mbox{Hollywood} provides
lots of examples of the dangers awaiting us
here. 
Such a law \morehighlight{would, of course, cause} problems
in running any sort of Turing test. However, 
\morehighlight{I expect
that the current discussion about replacements for 
the Turing test will eventually move from tests for AI
based on deception to tests that quantify
explicit skills and intelligence.} 
\morehighlight{Some related legislation has been
put into law for guns. In particular, 
Governor Schwarzenegger signed legislation in
September 2004 that prohibits
the public display of toy guns in California 
unless they are clear 
or painted a bright color to differentiate them from real firearms. 
The purpose of this law is to prevent police officers 
mistaking toy guns for real ones. 
}

The second part of the law states that
\highlight{autonomous systems} need to identify
themselves at the start of any interaction
with another agent. Note that this other agent
might even be another AI. This is
intentional. If you send you AI bot out
to negotiate the purchase of a new car,
you want the bot also to know whether it
is dealing with a dealer bot or a person.
You wouldn't want the dealer bot to be
able to pretend to be a human just because is
was interacting with your bot. 
The second part of the law is designed to
reduce the chance that autonomous systems 
are accidently mistaken for what they are not. 

Let's consider \highlight{
four up and coming areas} where
this law might have bite. First, consider autonomous vehicles. 
I find it a real oversight that the first
piece of legislation that permits autonomous
vehciles on roads, the AB 511 act in 
Nevada, says nothing at all about such
vehicles being identified to other road users
as autonomous. 
A Turing Red Flag law, on the other
hand, would require
an autonomous vehicle identify itself as autonomously
driven both to human drivers and to other
autonomous vehicles. 
\highlight{There are many situations where
it could be important to know that another road
vehicle is being driven autonomously. For example,
when a light changes we can suppose that an 
autonomous vehicle approaching the light will
indeed stop, and so save us
from having to brake hard to avoid an
accident. As a second example, 
if an autonomous car is driving in 
front of us in fog, we can suppose
it can see a clear road ahead using its
radar. For this reason, we do not have to leave
a larger gap \morehighlight{in case it has to brake} suddenly. 
\morehighlight{As a third example, at a four way intersection,
we can suppose an autonomous car will not
aggressively pull out when it does not have
right of way.} And as a fourth and final
example, if an autonomous car arrives
at a diversion, we might expect
it to drive more slowly as it tries
to work out where the road is now going. 
}

How should an autonomous
vehicle identify itself? I don't suppose this
should be with a person walking in front
with a red flag. This was too restrictive
even back in 1865. Autonomous vehicles might
have to carry distinctive plates, just like
we require learner drivers to identify themselves
on the roads today. Or autonomous vehicles might have to 
display a magenta flashing light
whenever they are being operated
autonomously. In addition, autonomous
vehicles should broadcast their location,
velocity and autonomy to neighbouring 
vehicles. 

In June 2015, Reuters reported
that two autonomous cars, one
from Google and the other from Delphi Automotive Plc
nearly had an accident on a Silicon Valley street. 
The Google car apparently cut off the Delphi
car as it was about to perform a lane change.
The Delphi car then had to take ``appropriate
action'' to avoid an accident. Clearly, 
it would help prevent such incidents if 
autonomous vehicles were required to
broadcast their location and intentions. 
Of course, adequate safeguards
will also need to be put in place that 
such broadcasting does not compromise the
privacy of their human occupants. 

Once autonomous vehicles are
commonplace, other motorists will
expect to meet autonomous vehicles. But
before this time, it will be important
to know that the oncoming vehicle
is somewhat special and may behave
differently to a human driver.
In the past, it was less necessary
for autonomous vehicles to identify
themselves. There were few of them and
most looked a little
odd. The presence of large rotating
LIDAR sensors on their roofs has tended to give the game
away. However, now that the technology
is shrinking and autonomous cars are starting to 
look more normal, it will become increasingly
difficult to know when a car is being
driven autonomously. 
There's also a reverse argument. Suppose
autonomous vehicles are (or become) better
drivers than humans. Then just like today
when we demand less experienced and
capable learner drivers are identified,
we might demand that human drivers are
identified apart from autonomous drivers.
A Turing Red Flag law will achieve this.

Let's move on to the second 
area where a Turing Red Flag law
might start having impact shortly.
Consider virtual assistants like
Apple's Siri or Microsoft's Cortana. Here's
a conversation I just had with Siri.

~ \\

\fbox{
\begin{minipage}[t]{0.42\textwidth}
Siri. {\em What can I help you with?}

Q. Are you a computer?

Siri. {\em I can neither confirm nor deny my
current existential status.}

Q. Are you a human?

Siri. {\em I'm an assistant. That's all that
matters. }

Q. Are you an AI?

Siri. {\em That's a rather personal question.}
\end{minipage}
}

~ \\

Based on conversations like these, it would
appear that Siri is coming close
to violating this proposed Turing Red Flag law.
It begins its conversations 
without identifying itself as a computer,
and it answers in a way that, \highlight{
depending on
your sense of humour}, might deceive. At least,
in a few years time, when the dialogue is
likely more sophisticated, you can imagine
being deceived. Of course, few if any people 
are currently deceived into believing that Siri is
human. It would only take a couple of questions
\morehighlight{for Siri to reveal that it is not human}. 
Nevertheless, it is a dangerous precedent
to have technology like this in
everyday use on millions of smartphones
pretending, albeit poorly, to be human. 

There are also several more trusting groups that
could already be deceived. My five year old
daughter has a doll that uses a bluetooth
connection to Siri to answer general questions.
I am not so sure she fully appreciates
that it is just a smartphone doing
all the clever work here. Another 
\morehighlight{troubling} group are patients with
Alzheimer's disease and other forms of dementia. 
Paro is a cuddly robot seal
that has been trialled as therapeutic 
tool to help such patients. Again, 
some people find it troubling that 
a robot seal can be mistaken for real. 
Imagine then how much more troubling
society is going to find it when such
patients mistake AI systems for humans? 

\highlight{
Let's move onto a third example, online poker.
This is a multi-billion dollar industry so
it is possible to say that the stakes are high.
Most, if not all, online poker sites
already ban computer bots from playing. 
Bots have a number of advantages,
certainly over weaker players.
They never tire. They can compute
odds very accurately. They can track
historical \morehighlight{play very accurately}. 
Of course, in the current state of the art, they
also have disadvantages \morehighlight{such as} understanding
the psychology of their opponents. 
Nevertheless, in the interest of fairness,
I suspect most human poker players would prefer
to know if any of their opponents
was not human. A similar argument
could be made for other online computer games.
You might want to know if you're being
``killed'' easily because your opponent
is a computer bot with lightning fast reflexes. 

I'll end with a fourth example, computer
generated text. \morehighlight{Associated Press 
now generates most
of its US corporate earnings reports
using a computer program developed
by Automated Insights \cite{apnews}}. A narrow interpretation 
might rule such computer generated
text outside the scope of a Turing Red Flag law. 
Text generation algorithms are typically 
not autonomous. Indeed, they are typically
not interactive. However, 
if we consider a longer time scale, 
then such algorithms are interacting in some 
way with the real world, and they may well be mistaken for
human generated text. Personally, I would prefer to 
know whether I was reading text written by 
human or computer. It is likely to impact on my
emotionally engagement with the text.
But I fully accept that we are now in 
a grey area. \morehighlight{You
might be happy for automatically
generated tables of stock prices and weather
maps to be unidentified as computer generated,
but perhaps you do want match reports to be identified as such? }
What if 
the commentary on the TV show
covering the World Cup Final
is not Messi, one of
the best footballers ever,
but a computer that just happens to 
sound like Messi?
\morehighlight{And should you be
informed if the beautiful
piano music being played on the radio is
composed by Chopin or by a computer
in the style of Chopin?}
These examples illustrate that we
still have some way to go working out
where to draw the line with any Turing Red Flag
law. But I would argue, there is
a line to be drawn somewhere here.
}

There are several arguments that can be raised against a Turing
Red Flag law. 
One argument is 
that it's way too early to be worrying about
this problem now. Indeed, by flagging 
this problem today, we're just adding to
the hype around AI systems breaking bad.
There are several reasons why I discount this argument. 
First, autonomous vehicles are likely only a few
years away. In June 2011, Nevada's Governor
signed into law AB 511, the first legislation 
anywhere in the world 
which explicitly permits autonomous vehicles. 
As I mentioned before, I find it surprising that the bill
says nothing about the need for
autonomous vehicles to identify themselves. 
\highlight{
In Germany, autonomous vehicles are
currently prohibited based on the
1968 Vienna Convention on Road Traffic to which Germany 
and 72 other countries follow. However, 
the German transport minister formed a 
committee in February 2015 to 
draw up the legal framework that would make
autonomous vehicles permissible on German roads.
This committee has been asked to present
a draft of the key points in such 
a framework before the Frankfurt car fair in September
2015. We may therefore already be running late to ensure
autonomous vehicles identify themselves on German roads. }
Second, many of us have already
been fooled by computers. Several
years ago a friend asked me how the 
self-service checkout could recognise
different fruit and vegetables. I 
hypothesised a classification algorithm,
based on colour and shape. But then
my friend pointed out the CCTV display
behind me with a human operator doing
the classification. The boundary between
machine and man is quickly blurring. 
Even experts in the field can be mistaken. 
A Turing Red Flag law will help
keep this boundary sharp. 
Third, humans are often quick to
assign computers with more 
capabilities than they actually
possess. The last example illustrates
this. As another example, I let some students play
with an Aibo robot dog, and they
quickly started to ascribe the Aibo
with emotions and feelings, \highlight{neither of
which the Aibo has. Autonomous systems
will be fooling us as human long 
before they actually are capable to
act like humans.} Fourth, one of the most dangerous times for
any new technological is when the technology
is first being adopted, and society has not 
yet adjusted to it. It may well be, \morehighlight{as
with motor cars today}, society decides
to repeal any Turing Red Flag laws once
AI systems become the norm. But whilst they
are rare, we might well choose to act
a little more cautiously. 

In many states of the USA, as well as many countries
of the world including Australia, Canada and Germany,
you must be informed if your telephone
conversation is about to be recorded. 
Perhaps in the future it will be 
routine to hear, {\em ``You are about
to interact with an AI bot. If 
you do not wish to do so, please
press 1 and a real person will come
on the line shortly.''}

\bibliographystyle{aaai}
\bibliography{/Users/twalsh/Documents/biblio/a-z,/Users/twalsh/Documents/biblio/a-z2,/Users/twalsh/Documents/biblio/pub,/Users/twalsh/Documents/biblio/pub2}

\end{document}